\documentclass[11pt,a4paper]{article}
\usepackage[hyperref]{emnlp-ijcnlp-2019}
\usepackage{times}
\usepackage{latexsym}

\usepackage{url}
\usepackage{graphicx}
\usepackage{pgfplots}
\usepackage{multirow}
\usepackage{amsmath}

\usepackage{algorithm}
\usepackage[noend]{algpseudocode}
\aclfinalcopy 

\title{
Linguistically Informed Relation Extraction and Neural Architectures for Nested Named Entity Recognition in BioNLP-OST 2019}

\newcommand*{\affaddr}[1]{#1} 
\newcommand*{\affmark}[1][*]{\textsuperscript{#1}}

\author{*Usama Yaseen\affmark[1,2], *Pankaj Gupta\affmark[1,2], Hinrich Sch\"{u}tze\affmark[2]\\ 
 \affaddr{\affmark[1]Corporate Technology, Machine-Intelligence (MIC-DE), Siemens AG  Munich, Germany}\\
  \affaddr{\affmark[2]CIS, University of Munich (LMU) Munich, Germany} \\
  {\tt \{usama.yaseen,pankaj.gupta\}@siemens.com}
}

\date{}

\begin{document}
\maketitle
\begin{abstract}

  Named Entity Recognition (NER) and Relation Extraction (RE) are essential tools in distilling knowledge from biomedical literature. This paper presents our findings from participating in BioNLP Shared Tasks 2019. We addressed Named Entity Recognition including nested entities extraction, Entity Normalization and Relation Extraction. Our proposed approach of Named Entities can be generalized to different languages and we have shown it's effectiveness for English and Spanish text. We investigated linguistic features, hybrid loss including ranking and Conditional Random Fields (CRF), multi-task objective and token-level ensembling strategy to improve NER. We employed dictionary based fuzzy and semantic search to perform Entity Normalization. Finally, our RE system employed Support Vector Machine (SVM) with linguistic features.
  
  Our NER submission (team:MIC-CIS) ranked first in {\it BB-2019 norm+NER task} with standard error rate (SER) of {\bf 0.7159} and showed competitive performance on {\it PharmaCo NER} task with F1-score of {\bf 0.8662}. Our RE system ranked first in the {\it SeeDev-binary Relation Extraction Task} with F1-score of {\bf 0.3738}.
  
\end{abstract}

\newcommand\blfootnote[1]{%
\begingroup
\renewcommand\thefootnote{}\footnote{#1}%
\addtocounter{footnote}{-1}%
\endgroup
}

\section{Introduction}

\blfootnote{* Equal Contribution}Extracting knowledge from scientific articles is a challenging but very important problem. This becomes especially critical for biomedical literature which is growing at an increasing rate of at least 4\% per year, as of June 2019 there are 30 Million documents in PubMed \cite{DBLP:journals/biodb/Lu11}. Named Entity Recognition (NER) \cite{DBLP:conf/bionlp/Settles04, gupta2016table, DBLP:conf/naacl/LampleBSKD16} in the context of biomedical domain refers to the task of identifying the name of the biological entities e.g. name of a bacteria. Relation extraction\footnote{Event extraction is treated as RE in this work} (RE) \cite{kambhatla2004combining, mcdonald2005simple, lever2016verse, DBLP:conf/naacl/GuptaRS18} refers to identifying relations among biological entities (binary or n-ary). 

\begin{figure}[t]
\centering
    \includegraphics[scale=0.75]{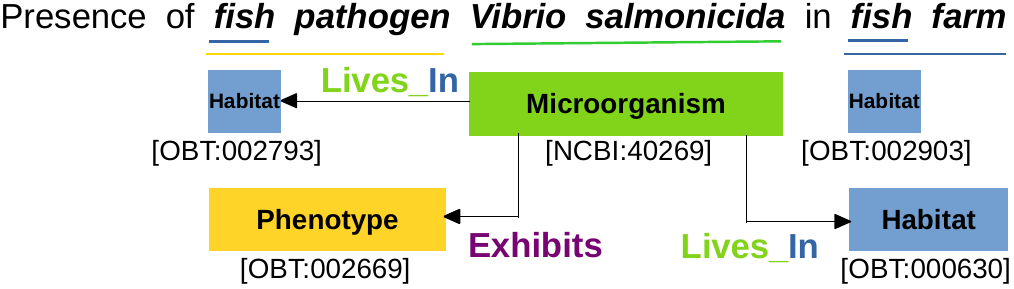}
    \caption{An illustration of (nested) NER + Normalization and Relation Extraction in Biomedical entities. Each rectangular box spans an entity, where the overlapping spans indicate nested entities. E.g., {\it fish} is a nested entity (a sub-concept) of type {\it Habitat} within the parent entity {\it fish pathogen} of type {\it Phenotype}. The identifiers (e.g. OBT:002669, NCBI:40269, etc.) refer to unique IDs in Biomedical databases (i.e., OBT $\rightarrow$ OntoBiotope Ontology and NCBI $\rightarrow$ NCBI Taxonomy), used to perform entity normalization (i.e., entity linking).  The arrows indicate binary relationships.   
}
    \label{fig:overview}
\end{figure}

Figure \ref{fig:overview} illustrates an example of (nested) NER and RE consisting of five entities, where three entities participate in two distinct relationships. It is often required to link named entity(s) to a unique reference in database(s).
For instance,  one of the two occurrences of {\it fish} refers to {\it marine fish} while the second refers to a {\it farm fish}, where the two entities are linked (or normalized) to different identifiers (e.g., OBT:002793 and OBT:002903) in the biomedical database (e.g., OntoBiotope Ontology).   
\begin{figure*}[t]
    \centering
    \includegraphics[scale=0.74]{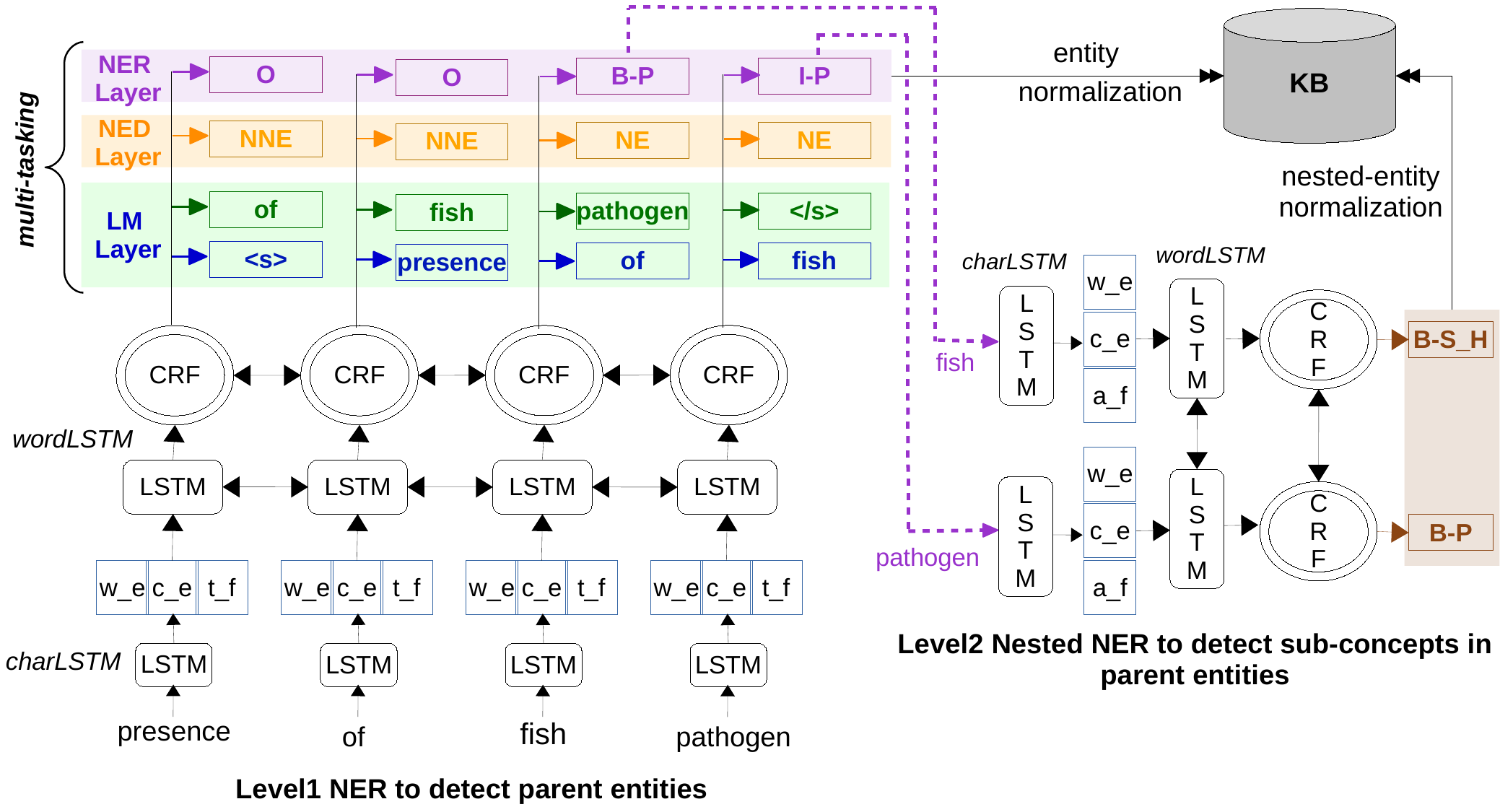}
    \caption{System Architecture for NER task, consisting of two bi-LSTM-CRF architectures: Level1 NER to detect parent entities and Level2 Nested NER to detect sub-concepts within the parent entities (output of Level1 NER).  Here,  \emph{w\_e}: a word embedding vector; \emph{c\texttt{\char`_}e}: an embedding vector for a word computed using character-level bi-directional LSTM;  \emph{t\texttt{\char`_}f}: a vector of additional linguistic features; B\texttt{\char`_}P: B\texttt{\char`_}Pathogen;  B-S\texttt{\char`_}H: a sub-concept of type \emph{Habitat} detected by the Level2 Nested NER run over the the parent entity.}
    \label{fig:task2_pipeline}
\end{figure*}
The act of linking entities to standard entities with a unique identifier is known as \emph{entity normalization} and is challenging as several entity mentions can correspond to the same standard entity (or unique identifier), e.g. {\it E. coli}, {\it Bacillus coli} and {\it Bacteriumcoli} refer to the standard entity \emph{Escherichia coli} in the database.  
The linking process relies on knowledge base (KB) search (heuristic OR semantic) in order to resolve entities. 

NER is a critical primitive step in the NLP pipeline as downstream tasks such as RE, text classification, Question Answering (QA) etc., depend on it. Even though several methods have been devised to engineer reliable NER systems; however, most of them don't explicitly address the extraction (or recognition) of nested entities, especially required in the biomedical domain. 
{\it Nested entity} is defined as an entity or sub-concept which is part of a longer entity (i.e., a parent).  
For instance in the Figure \ref{fig:overview}, \emph{fish} is a nested entity as it is part of a parent entity  \emph{fish pathogen}. In this work, we have also investigated extracting nested entities via two bi-LSTM-CRF \cite{DBLP:conf/naacl/LampleBSKD16} networks: one for parent detection and another for nested entities with the parent entity. 

\section{Task Description and Contribution}
We participate in the following three tasks organized by BioNLP workshop 2019:
(1) {\bf PharmaCoNER}:  Recognition of pharmaceutical drugs and chemical entities in Spanish text.  
(2) {\bf BB-norm+NER}: Recognition of {\it Microorganism}, {\it Habitat} and {\it Phenotype} entities and normalization with NCBI Taxonomy and OntoBiotope habitat concepts.
(3) {\bf SeeDev Binary RE}: Binary Relation extraction of genetic and molecular mechanisms involved in plant seed development. 

Following are our multi-fold {\it contributions}:
\begin{enumerate}
\item To address NER tasks, we have employed neural network based sequence classifier, i.e., bi-LSTM-CRF and investigated multi-tasking of named entity detection (NED) and language modeling (LM). We further introduced hybrid loss including CRF and ranking. We also incorporated linguistic features such as POS, orthographic features, etc.  We apply the proposed modeling approaches to both English and Spanish texts. 
Comparing with other systems, our submission (Team: MIC-CIS) is ranked $1^{st}$ in {\it BB-norm+NER} task \cite{bossy2019} with standard error rate of $0.7159$. In {\it PharmaCoNER} task \cite{pharmaconer2019}, our submission scored F1-score of $0.8662$.

\item To address RE task, we employed linguistic and entity features in SVM.   Our submission (Team: MIC-CIS) is ranked 
 $1^{st}$ in SeeDev-binary RE task \cite{ChaixDFVBBDZBLN16} with F1-score of $0.3738$. 
 \end{enumerate}

The code to reproduce our results is available at: \url{https://github.com/uyaseen/bionlp-ost-2019}.
\section{Methodology}

In the following sections we discuss our proposed model for NER and RE.

\begin{table}[t]
\centering
\resizebox{.49\textwidth}{!}{
\begin{tabular}{l | l}
\multicolumn{1}{c|}{\textbf{Features}} & \multicolumn{1}{c}{\textbf{Description}} \\ \hline
word-cap & capitalization features \\ \hline
POS & parts-of-speech tags \\ \hline
\multirow{2}{*}{ortho} & orthographic features \\ 
& e.g. \emph{Egg Pulp, 97} encoded as \emph{Ccc Ccccp nn} \\ \hline
tri-gram & tri-gram as features \\ \hline
five-gram & five-gram as features \\ \hline
length & length of the word \\ \hline
sdp-rel & dependency relation tag \\ \hline
\multirow{2}{*}{alpha-features} & detect if certain linguistic pattern occurred \\
& in the current word or the next word \\ 
\end{tabular}
}
\caption{Word-level features for NER. The features are encoded as embeddings, except the \emph{alpha} features that are represented as one-hot vector.}
\label{ner_features}
\end{table}

\subsection{Neural Architectures for NER}

Figure \ref{fig:task2_pipeline} describes the architecture of our model, where we design two sequence taggers {\it Level1 NER} and {\it Level2 Nested NER} to extract parent and nested entities respectively. Furthermore, Level1 NER can be configured in two modes: 
(1) LSTM-CRF \cite{DBLP:conf/naacl/LampleBSKD16} with word embeddings ($w\_e$), character embeddings ($c\_e$)  and token-level features ($t\_f$) such as POS, capitalization features, word shape, etc. (refer to table \ref{ner_features} for the complete list of word level features)
(2) {\it LSTM-CRF+Multi-task} that performs entity detection and language modelling as auxiliary tasks.
Note that Level2Nested NER only  operates on the parent entities detected by Level1 NER. The parent and nested entities are than normalized to unique identifiers in KB by our entity normalization algorithm.

\subsubsection{BiLSTM-CRF}
The input to LSTM is a sequence of word features $(\mathbf{w}_1, \mathbf{w}_2, \ldots, \mathbf{w}_n)$ and they compute a hidden state for each element in the sequence $(\mathbf{h}_1, \mathbf{h}_2, \ldots, \mathbf{h}_n)$. This hidden state can be used to jointly model tagging decisions using CRF \cite{Lafferty2001ConditionalRF}. CRF imposes ordering constraints on the tagging decisions e.g. I\texttt{\char`_}Habitat should always be preceded by B\texttt{\char`_}Habitat.
For an input sentence,
$$\mathbf{W} = (\mathbf{w}_1, \mathbf{w}_2, \ldots, \mathbf{w}_n),$$
we consider a matrix $\mathbf{P}$ of scores output by the bidirectional LSTM. The size of $\mathbf{P}$ is $n~\times~k$, where $k$ is the 
number of distinct tags, and $P_{i, j}$ corresponds to the score of the $j^{th}$ tag of the $i^{th}$ word in a sentence. 
For a sequence of predictions
$$\mathbf{y} = (y_1, y_2, \ldots, y_n),$$
we define its score to be
$$s(\mathbf{X}, \mathbf{y})=\sum_{i=0}^{n} A_{y_i, y_{i+1}} + \sum_{i=1}^{n} P_{i, y_i}$$
where the matrix $\mathbf{A}$ express transition scores such that $A_{i, j}$ represents the score of a transition from the tag $i$ to tag $j$. We add \textit{start} and \textit{end} tag to the set of possible tags, therefore, the size of A is $k+2$.
During training, we minimize the negative log-probability of the correct tag sequence:
\begin{align}
\log(p(\mathbf{y} | \mathbf{X})) &= s(\mathbf{X}, \mathbf{y}) - \log \left( \sum_{\mathbf{\widetilde{y}} \in \mathbf{Y_X}} e^{s(\mathbf{X}, \mathbf{\widetilde{y}})} \right) \nonumber \\
&= s(\mathbf{X}, \mathbf{y}) - \underset{{\mathbf{\widetilde{y}} \in \mathbf{Y_X}}}{logadd}\ s(\mathbf{X}, \mathbf{\widetilde{y}}), 
\end{align}
\begin{align}
loss_{CRF} = - \log(p(\mathbf{y} | \mathbf{X}))
\end{align}

\subsubsection{Hybrid Loss: CRF + Ranking}

We use a variant of ranking loss function proposed by \newcite{Sa:15}. Ranking maximizes the distance between the true label $y^+$ and the most competitive label $c^-$:

$ loss_{ranking} = max(0, 1 + (\gamma * (m^+ - y^+)) + (\gamma * (m^- + c^-))$

where $\gamma$ is the scaling factor that penalizes the predictions, $m^+$ and $m^-$ are margins for correct and incorrect labels respectively. We follow \newcite{DBLP:conf/icassp/VuGAS16} to set the values of margins. 

The hybrid loss function hence is the sum of CRF tagging loss and ranking loss:
\[loss_{hybrid} = loss_{CRF} + \alpha \ \cdot \   loss_{ranking}\]
where $\alpha \in [0,1]$, weighs the contribution of ranking loss in the overall loss value.
During training we minimize the hybrid loss and found it to improve the F1 score for both {\it BB-norm+NER} and {\it PharmaCoNER tasks}.

\subsubsection{Multi-Tasking of Named Entity Recognition, Detection and Language Modelling}
We employed auxiliary objectives of named-entity detection (NED) \cite{Ag:19} and bidirectional language modelling (LM) \cite{re:17} in our model. Usually these auxiliary objectives acts as regularizes \cite{Co:08} and improves the overall performance. With these multi-tasking objectives, for each word token our model predicts the NED tag, next word, previous word and the NER tag\footnote{we used IOBES tagging scheme}. LM and NED layers in figure \ref{fig:task2_pipeline} realizes NED and LM objectives respectively. Note that Multi-tasking is only enabled at train time and requires no additional labelling.

\begin{table*}[t]
\centering
\resizebox{.90\textwidth}{!}{
\begin{tabular}{c | l || c | l}
\textbf{General Features} & \textbf{Description} & \textbf{Entity Features} & \textbf{Description} \\ \hline
\multirow{2}{*}{bow} & bag-of-words (bow) representation & \multirow{2}{*}{entity-pos} & position of entity in the bow \\ & of the complete sentence & & representation \\ \hline
\multirow{4}{*}{bow-partial} & bow representation of the between & & \\ & context (i.e. word tokens between & \multirow{2}{*}{entity-type} &  \multirow{2}{*}{type of the entity mentions} \\ & target entities) including three & &  \\ & words to the target entities  & & \\ \hline
\multirow{2}{*}{bow-lemma} & bow representation of the lemmatized & \multirow{2}{*}{dist-entities-cat} & distance between target entities \\ & tokens in the between context & & as categorical \\ \hline
pos-tags & part-of-speech tags & dist-entities & distance between target entities \\ \hline
sdp & shortest dependency path as bow & entity-count & count of entities in between context \\ \hline
\multirow{2}{*}{sdp-len} & length of shortest dependency path & \multirow{2}{*}{entity-count-cat} & count of entities in between context \\ & as scalar & & as categorical \\ \hline
sdp-rel & dependency relation tag & e1 type = e2 type & if type of e1 and e2 is same \\ \hline
emb-sdp & average embeddings of sdp & sdp-entity & sdp with entity as bow \\ \hline
\multirow{2}{*}{keyword-vec} & if current word is part of feature & \multirow{2}{*}{entity-patterns} & check if certain linguistic patterns \\ & list of relations & & occur in the vicinity of target entities \\
\end{tabular}
}
\caption{General and Entity features used in Relation Extraction}
\label{relation-extraction-features}
\end{table*}

\subsubsection{Nested Entities}
The dataset of \emph{BB-norm+NER} task contains $17.4\%$ nested entities \footnote{\url{https://groups.google.com/d/msg/bb-2019/A2MuFYiPQIY/9YtMmakeBQAJ}} which cannot be extracted by standard Bi-LSTM CRF model. We employed
two Bi-LSTM-CRF models: {\textit Level1 NER} model to detect parent entities and {\textit Level2 Nested NER} model to detect nested entities. Figure \ref{fig:task2_pipeline} (right) shows the architecture of Level2 Nested NER. The parent entities detected by Level1 NER are fed to Level2 Nested NER to detect nested entities in the parent entities. Level2 Nested NER has the same architecture as Level1 NER but without the multi-tasking objectives.
It is easy to see that current architecture can only detect nested entities at level 2. The final output of model is the aggregation of parent entities and nested entities.

\subsubsection{Entity Normalization}

\begin{algorithm}[t]
\caption{{\small 
Entity Normalization}}\label{algo:entity-norm} 
\small
{ 
\begin{algorithmic}[1]
\Statex \textbf{Input}: NE, NE\_Type
\Statex \textbf{Output}: RF\_ID
\Statex \textbf{Output}: NE\_PRED (Optional)
\State RF\_ID = None
\State IF NE\_Type == 'Microorganism':
\State \quad found, RF\_ID = exact\_match(NE, NCBI)
\State \quad if not found:
\State \quad \quad found, RF\_ID = fuzzy\_match(NE, NCBI)
\State \quad return RF\_ID
\State ELSE
\State \quad found, RF\_ID = exact\_match(NE, NCBI)
\State \quad if not found:
\State \quad \quad found, RF\_ID = fuzzy\_match(NE, NCBI)
\State \quad if found:
\State \quad \quad \# LABEL UPDATE !
\State \quad \quad NE\_PRED = 'Microorganism'
\State \quad \quad return RF\_ID, NE\_PRED
\State \quad found, RF\_ID = exact\_match(NE, OBT)
\State \quad if not found:
\State \quad \quad found, RF\_ID = semantic\_search(NE, OBT)
\State return RF\_ID
\end{algorithmic}}
\end{algorithm}

The goal of entity normalization (entity linking) is to map noisy predicted entities in text to canonical entities in knowledge base (KB). This is challenging because: (1) not all variations of textual forms for a canonical entity exists in the KB, (2) syntactic variations in the predicted entity mentions due to misspellings, abbreviations, acronyms and boundary errors. 

For {\it BB-norm+NER} task, we used two Biomedical databases {\it OntoBiotope Ontology} and {\it NCBI Taxonomy}. OntoBiotope Ontology contains $3,602$ canonical forms of type \emph{Habitat} and \emph{Phenotype}. NCBI Taxonomy contains $1,082,401$ records for type {\it Microorganism}. We employed exact, fuzzy and semantic (embedding) search to perform entity normalization. Algorithm $1$ illustrates the detailed steps of our algorithm, note that type and order of search depends on the predicted named entity type. We also employed {\it caching} to minimize pairwise comparisons and improve the overall run-time efficiency.

\begin{table}[t]
\center
\renewcommand*{\arraystretch}{1.1}
\resizebox{.495\textwidth}{!}{
\begin{tabular}{r | c | ccc | c | c}
&\multirow{2}{*}{\bf Tokens} & \multicolumn{3}{c|}{\bf Models} & \textbf{Voting} & \textbf{Post-} \\
 & & \textit{M1} & \textit{M2} & \textit{M3} &  & \textbf{processing} \\ \hline
r1 & Presence & O & O & O & O & O \\ \hline
r2 & of & O & O & B-H & O & O \\ \hline
r3 & fish & I-H & B-H & I-H & I-H & B-H \\ \hline
r4 & pathogen & I-H & I-P & I-P & I-P & B-P \\ \hline
r5 & Vibrio & B-M & B-M & B-M & B-M & B-M \\ \hline
r6 & salmonicida & I-M & O & I-M & I-M & I-M \\ \hline
r7 & in & B-H & O & O & O & O \\ \hline
r8 & fish & B-H & O & B-H & B-H & B-H \\ \hline
r9 & farm & I-H & O & I-M & I-H & I-H \\ \hline
r10 & . & O & O & O & O & O 
\end{tabular}}
\caption{NER: Ensembling and Post-processing correcting individual models mistakes. Here, B, P and M refer to Habitat, Phenotype and Microorganism, respectively.}
\label{ensemble_and_post_processing}
\end{table}

\subsubsection{Post-processing for NER+norm}
Our model (see Figure \ref{fig:task2_pipeline}) employs CRF at decoding step to impose boundary ordering constraints on the predicted named entity types e.g. \emph{I} should always be preceded by a \emph{B} token. But our model does not always respect such ordering constraints and therefore, we resolve boundary inconsistencies at inference time to make the NER labels consistent. \emph{Post-processing} column in the Table \ref{ensemble_and_post_processing} illustrates the post-processing resolving inconsistent labels after the voting on majority labels, consider row {\it r3} where post-processing correctly imposes the semantics of boundary ordering by changing \emph{I-Habitat} to \emph{B-Habitat}.

\subsection{Relation Extraction}

Deep Learning based methods are state of the art in relation extraction \cite{WuRE2019,wang-etal-2016-relation} but they require large amount of labelled training data. In cases when enormous training data is not available than Kernel methods like Support Vector Machines (SVM) are an optimal choice. We employed SVM for performing relation extraction. One of the downsides of SVM is that they usually require lots of hand-crafted features to train properly. Table \ref{relation-extraction-features} lists computed general and entity features.

Our best model was trained with Radial Basis Function (RBF) Kernel with value of penalty parameter $C$ determined by grid search for each dataset. We employed oversampling and class-weight penalization to handle imbalanced data. Surprisingly oversampling did not provide any performance improvement therefore, final models were trained only with higher class weights for minority classes. We did not normalize any input feature as it resulted in reduced performance.

In relation extraction participating entities are not known in advance, the usual practise is to test every valid pair of entities for a relation. We employed heuristic of {\it token counts} between entities to filter the probable invalid relations. The value of token counts was determined using cross-validation. 
\subsection{Ensemble Strategy}

Bagging is a helpful technique to reduce variance without impacting bias of the learning algorithm. We employed a variant of \emph{Bagging} \cite{Br:96} which makes sure that every sample in the training set is part of the development set at least once and vice versa. We created three data folds and trained the model using optimal configuration on each fold, prediction on test involves majority voting among the \emph{three} trained models.

The commonly used tagging schemes (BIO, BIOES etc.,) for NER contains information about the boundary of an entity along with the class of an entity, which is spitted by the model at each time-step. Due to this dual information in a single output, maximum voting is not trivial as models can not only disagree on the class but also on the boundary of an entity. Empirically we found that our model is better at predicting the class of an entity rather than the boundary of an entity, therefore, we followed the strategy \emph{class determines the boundary}. In cases when voting results in a tie, we take the prediction of the \emph{confident} model, we treat the model trained on original train/dev split as the confident model. We also experimented with an extreme version of ensembling where we aggregate the output of every model with distinct spans, as expected this improves the recall but with the cost of reduced precision. One possible optimization to this ensemble strategy is to only aggregate the non-overlapping spans to control reduction in precision without much decrease in recall, we will explore this as a future work.
Table \ref{ensemble_and_post_processing} shows the ensemble correcting individual model's erroneous predictions.

In case of ensemble for RE, we followed the straight forward approach of majority voting at sentence level for each test sample.
\label{section:bagging}

\begin{table}[t]
\center
\renewcommand*{\arraystretch}{1.2}
\resizebox{.46\textwidth}{!}{
\begin{tabular}{c | c | c | c}
\textbf{Task} & \textbf{Train} & \textbf{Dev} & \textbf{Test} \\ \hline
\multicolumn{4}{c}{\textbf{Sentence Counts}}  \\ \hline
PharmaCo & 8068 & 3748 & 3930 \\ \hline
SeeDev  & 644 & 308 & 466 \\ \hline
BB-norm+ner  & 822 & 413 & 735 \\ \hline
\multicolumn{4}{c}{\textbf{PharmaCoNER Entities}}  \\ \hline
NORMALIZABLES & 2304 & 1121 & 859 \\ \hline
PROTEINAS & 1405 & 745 & 973 \\ \hline
UNCLEAR & 89 & 44 & 34 \\ \hline
NO\_NORMALIZABLES & 24 & 16 & 10 \\ \hline
\multicolumn{4}{c}{\textbf{BB-norm+NER Entities}}  \\ \hline
Habitat & 1118 & 610 & - \\ \hline
Microorganism & 739 & 402 & - \\ \hline
Phenotype & 369 & 161 & - 
\end{tabular}
}
\caption{Dataset statistics for NER}
\label{table:ner-statistics}
\end{table}

\section{Experiments and Results}

\subsection{Dataset and Experimental Setup}

\textbf{Data:} We employed bagging (discussed in section \ref{section:bagging}) to split the annotated corpus into 3-folds. We used pre-processed versions of datasets for BB-norm+NER\footnote{\url{https://sites.google.com/view/seedev2019/supporting-resources}} and SeeDev\footnote{\url{https://sites.google.com/view/bb-2019/supporting-resources}} provided by the organizers. This pre-processed version comes with sentence splitting, word tokenization and POS tagging.

\textit{PharmaCoNER:} The dataset consists of four entity types with very few mentions of type \emph{UNCLEAR} and \emph{NO\_NORMALIZABLES} as shown in table \ref{table:ner-statistics}. Entities of type \emph{UNCLEAR} are ignored in the evaluation of this shared task but we still treat them as regular entities.

\textit{BB-norm+NER:} The dataset consists of three entity types with few mentions of type \textit{Phenotype} (see table \ref{table:ner-statistics}). The dataset also contains $3.6\%$ \textit{disconnected entities}\footnote{\url{https://groups.google.com/d/msg/bb-2019/A2MuFYiPQIY/9YtMmakeBQAJ}}, we did not employ any strategy to handle disconnected entities and instead treat them as separate (regular) entities.

\textit{SeeDev:} The dataset consists of $22$ binary relations among $16$ entity types. The dataset is highly imbalanced with zero instances of type \emph{Regulates\_Molecule\_Activity} and \emph{Composes\_Protein\_Complex} in the default development set.

\begin{table}[t]
\centering
\resizebox{.44\textwidth}{!}{
\begin{tabular}{c | c}
\hline
\textbf{Hyper-parameter} & \textbf{Value} \\ \hline
\multicolumn{2}{c}{{\bf NER}} \\ \hline
learning rate & 0.005 \\ \hline
character (char) dimension & 25 \\ \hline
hidden unit::char LSTM & 25 \\ \hline
POS dimensions & $25^{*}$, $50^{+}$ \\ \hline
Ortho dimension & $25^{*}$, $50^{+}$ \\ \hline
hidden unit::word LSTM & $200^{*}$, $100^{+}$ \\ \hline
word embeddings dimension & $200^{*}$, $100^{+}$ \\ \hline
length dimension & 10 \\ \hline
sdp\_rel & 10 \\ \hline
alpha\_features & 2 \\ \hline
ranking loss::$\alpha$ & 1.0 \\ \hline
ranking loss::$\gamma$ & 1.0 \\ \hline \hline
\multicolumn{2}{c}{{\bf RE}} \\ \hline
kernel & RBF \\ \hline
class-weights & $10.0$ \\ \hline
\end{tabular}
}
\caption{Hyper parameter settings for NER and RE. * and + denote the optimal parameters for BB-norm+ner and PharmaCoNER respectively.}
\label{table:hyper-params}
\end{table}

\textbf{Experimental Setup:} We found sub-word information to be very helpful in identifying entities and relations in biomedical domain and all our experiments used word embeddings trained using FastText \cite{Bo:16}. For tasks in English language we used FastText embeddings trained on PubMed \cite{Zh:19}. We don't employ any strategy for handling imbalanced classes for NER but have used class weighting by a factor of $10$ for all positive classes for RE. Table \ref{table:hyper-params} lists the best configuration of hyper-parameters for all the tasks.

\textit{PharmaCoNER:} We used \emph{SPACCC\_POS-TAGGER} \cite{soares2019} for sentence splitting, word tokenization and POS tagging. We trained FastText embeddings on the following corpora: IBECS \cite{Rodrguez2002TheSB}, IULA-Spanish-English-Corpus \cite{Marimon2017AnnotationON}, MedlinePlus \cite{Miller2000MEDLINEplusBA}, PubMed \cite{DBLP:journals/biodb/Lu11}, ScIELO \cite{Goldenberg2007S} and PharmaCoNer \cite{pharmaconer2019}. We trained embeddings on two variants of corpora: (1) Include train and development set of PharmaCoNER (2) Include complete dataset of PharmaCoNER. We concatenated these two embeddings to provide complementary information and found them to empirically work better than the embeddings trained on individual corpora variant. We compute micro-F1 using the script provided by the organizers on the dev set\footnote{\url{https://github.com/PlanTL-SANIDAD/PharmaCoNER-CODALAB-Evaluation-Script}}.

\textit{BB-norm+NER:} For training NER model we compute macro-F1\footnote{evaluation measure with strict boundary detection} \cite{TsaiWCLHHSH06} on the dev set. NER and Entity normalization together are evaluated using {\it Standard Error Rate (SER)} \cite{Bossy2015OverviewOT}. During the entity normalization step, the fuzzy and semantic search can resolve an entity mention to multiple normalization identifiers. Our algorithm returns top $5$ matched identifiers, however, we empirically found selecting the top most identifier gives superior performance.

\textit{SeeDev:} We adopted two strategies to create negative relation instances for train and dev+test set: (1){\it Train:} only consider sentences not participating in any positive relation
(2) {\it Dev+Test:} consider all the sentences. Negative relation instances are always created only among the valid combination of entity types. We also employed an extended version of keywords match of \newcite{Li:16} as a feature (referred as keyword vectors in table \ref{relation-extraction-features}).

\subsection{Results on Development Set}

\begin{table}[t]
\center
\renewcommand*{\arraystretch}{1.2}
\resizebox{.495\textwidth}{!}{
\begin{tabular}{r | l | c c c || c  c  c  c}
& \multirow{2}{*}{\textbf{Configration}} & \multicolumn{3}{c||}{PharmaCoNER} & \multicolumn{4}{c}{BB-norm+NER} \\
& & \textbf{P} & \textbf{R} & \textbf{F1} & \textbf{P} & \textbf{R} & \textbf{F1} & \textbf{SER} \\ \hline
& & \multicolumn{3}{c||}{\bf Fold=1} & \multicolumn{4}{c}{\bf Fold=1} \\
r1 & {\it BiLSTM-CRF}  & .884 & .773 & .824 & .809 & .474 & .598  & .576 \\
r2 & {\it + word-emb} & .892 & .857 & .874 & .831 & .526 & .644  & .524 \\ 
r3 & {\it + ortho} & .909 & .846 & .877 & .823 & .515 & .633  & .533 \\ 
r4 & {\it + POS} & .906 & .851 & .877 & .827 & .523 & .641  & .526 \\ 
r5 & {\it + multi-task} & .907 & .851 & .878  & .806 & .528 & .638  & .531 \\ 
r6 & {\it + length} & - & - & - & \textbf{.842} & .487 & .617  & .545  \\
r7 & {\it + ranking} & \textbf{.912} & \textbf{.860} & \textbf{.885} & .827 & .535 & {.650}  & {.520} \\ 
r8 & {\it + search} & - & - & - & .810 & {\bf .600} & {\bf .690} & {\bf .489} \\ \hline 
& & \multicolumn{3}{c||}{\textbf{Fold=2}} & \multicolumn{4}{c}{\textbf{Fold=2}} \\ 
r9 & {\it BiLSTM-CRF}  & .915 & .890 & .902 & .630 & .400 & .489 & - \\
r10 & {\it all features} & \textbf{.934} & \textbf{.889} & \textbf{.911} & \textbf{.719} & \textbf{.513} & \textbf{.599} & - \\ \hline
& & \multicolumn{3}{c||}{\textbf{Fold=3}} & \multicolumn{4}{c}{\textbf{Fold=3}} \\ 
r11 & {\it BiLSTM-CRF}  & .899 & .873 & .886 & .784 & .699 & .739 & - \\
r12 & {\it all features} & \textbf{.917} & \textbf{.877} & \textbf{.896} & \textbf{.813} & \textbf{.764} & \textbf{.788} & - \\

\end{tabular}
}
\caption{Scores on dev set using different features on {\it PharmaCoNER} and {\it BB-norm+NER} tasks. Here, + signifies feature accumulation to the last row.}
\label{table:ner-xptrs-dev}
\end{table}

To investigate the impact of features we incrementally enabled them and observe the affect on performance on dev set.

{\bf NER:} Table \ref{table:ner-xptrs-dev} shows the score on dev set for {\it PharmaCoNER} and {\it BB-norm+NER}. Observe that {\it FastText} embeddings (row r2) outperform randomly initialized embeddings (row r1) and contribute to biggest performance boost for both datasets.
Subsequently, {\it Orthographic} (row r3) and {\it POS} (row r4) features\footnote{Additionally, we  have employed document-topic proportion from neural topic models \cite{DBLP:conf/aaai/GuptaCBS19}, however, no significant gains were observed.} improve the scores for PharmaCoNER but surprisingly lower the score for BB-norm+NER. In row r5, we perform multi-tasking with auxiliary task of NED leading to improvement only for PharmaCoNER. Next, we incorporate hybrid loss including ranking (row r7) which consistently improves the score on both datasets. In row r8, we employed Brute Force Search (discussed in section \ref{section:analysis-on-dev}) that significantly reduce SER for BB-norm+NER.
Finally, we create an ensemble of (r7, r10, r12) and (r8, r10, r12) on test set for PharmaCoNER and BB-norm+NER respectively.

\begin{table}[t]
\centering
\resizebox{.45\textwidth}{!}{
\begin{tabular}{r | l | c | c | c}
& \textbf{Features} & \textbf{P} & \textbf{R} & \textbf{F1} \\ \hline
r1 & {\it bow-between}  & .0 & .0 & .0 \\ 
r2 &  {\it + class-weights} & .214 & .196 & .205 \\ 
r3 &  {\it + entity-type} & .157 & .589 & .248 \\ 
r4 &  {\it + sdp-entity} & .204 & .540 & .296 \\ 
r5 &  {\it + emb-sdp} & .212 & .479 & .294 \\ 
r6 &  {\it + lemma} & {\bf .220} & {\bf .478} & {\bf .301}
\end{tabular}
}
\caption{Scores on dev set using different features on {\it SeeDev} task. Here, + signifies feature accumulation to the last row.}
\label{table:re-xptrs-dev}
\end{table}

{\bf RE:} Table \ref{table:re-xptrs-dev} shows the score on dev set for {\it SeeDev}\footnote{Results are only reported for standard data fold as it was not trivial to change evaluation script for non-standard folds.}. In row r1, negative instances dominate the training set resulting in no learning. Observe that introduction of class weights (row r2) compensate the dominance of negative instances leading to F1 score of $0.205$. Next, we added {\it entity-type} (row r3) and {\it sdp-entity} (row 4) features, both of these features significantly improves F1 score i.e. by an absolute value of more than $4.0$. Subsequently, {\it emb-sdp} (row r5) and {\it lemma} (row r6) contribute to incremental improvements. Finally, we create an ensemble of {\it row r6} on all three data folds.

\subsection{Analysis on Development Set}

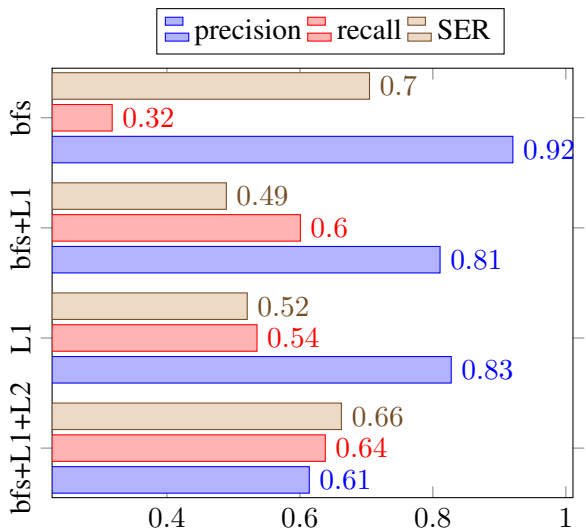
\begin{figure}
    \centering
    
\begin{tikzpicture}
\begin{axis}[
    xbar,
    enlargelimits=0.15,
    legend style={at={(0.5,-0.15)},
      anchor=north,legend columns=-1},
    xlabel={},
    symbolic y coords={bfs+L1+L2,L1,bfs+L1, bfs},
    ytick=data,
    yticklabel style={rotate=90},
    nodes near coords,
    nodes near coords align={horizontal},
    legend style={at={(0.2,1.14)},anchor=north west}
    ]
\addplot coordinates {(0.9203,bfs) (0.8276,L1) (0.8108,bfs+L1) (0.6141,bfs+L1+L2)};
\addplot coordinates {(0.3177,bfs) (0.5355,L1) (0.6007,bfs+L1) (0.6382,bfs+L1+L2)};
\addplot coordinates {(0.7047,bfs) (0.5208,L1) (0.48934,bfs+L1) (0.6624,bfs+L1+L2)};
\legend{precision,recall,SER}
\end{axis}
\label{figure-bb_ner-analysis}
\end{tikzpicture}
\caption{{\it BB-norm+NER}: Impact of brute-force search, Level1 NER and their aggregation on SER. Here bfs, L1 and L2 refer to {\it brute-force search}, {\it Level1 NER} and {\it Level2 Nested NER} respectively.}
\label{fig:analysis-bb-norm-ner}
\end{figure}

\textbf{BB-norm+NER:} We also explored approaching the problem of NER and entity normalization in a reverse manner by matching every entity mention from the biomedical databases (i.e. {\it NCBI Taxonomy} and {\it Ontobiotope}) in every sentence. This matching is indeed exhaustive search, we refer to it as {\it Brute-force search}.
Figure \ref{fig:analysis-bb-norm-ner}
shows the comparison of: (1) {\it brute-force search} (2) {\it Level1 NER} (3) aggregation of {\it brute-force search} and  {\it Level1 NER} (4) aggregation of {\it brute-force search}, {\it Level1 NER} and {\it Level2 NER}. Brute-force search yields high precision but a moderately low recall with SER value of $0.7$. In comparison, Level1 NER has significantly higher recall with a little reduction in precision yielding SER value of $0.52$. The aggregation of brute-force search and Level1 NER improves recall and lowers SER value to $0.49$. Finally, aggregation of brute-force search, Level1 NER and Level2 NER results in a balanced precision and recall values but an overall higher value of SER. Our submission on test set employed aggregation of {\it brute-force search} and {\it Level1 NER}. 

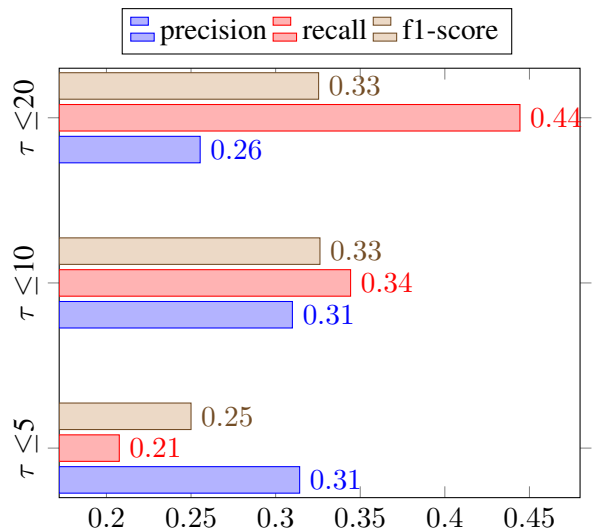
\begin{figure}
    \centering
    
\begin{tikzpicture}
\begin{axis}[
    xbar,
    enlargelimits=0.15,
    legend style={at={(0.5,-0.15)},
      anchor=north,legend columns=-1},
    ylabel={},
    symbolic y coords={$\tau\leq$5,$\tau\leq$10,$\tau\leq$20},
    ytick=data,
    yticklabel style={rotate=90},
    nodes near coords,
    nodes near coords align={horizontal},
    legend style={at={(0.12,1.14)},anchor=north west}
    ]
\addplot coordinates {(0.3142,$\tau\leq$5) (0.3099,$\tau\leq$10) (0.2554,$\tau\leq$20)};
\addplot coordinates {(0.2076,$\tau\leq$5) (0.3443,$\tau\leq$10) (0.4444,$\tau\leq$20)};
\addplot coordinates {(0.2500,$\tau\leq$5) (0.3262,$\tau\leq$10) (0.3254,$\tau\leq$20)};
\legend{precision,recall,f1-score}
\end{axis}
\end{tikzpicture}

\caption{{\it SeeDev}: Impact of 'token counts between target entities' heuristic on system performance.}
\label{fig:analysis-see-dev}
\end{figure}

{\bf SeeDev:} We employed the heuristic of token counts between target entities to filter potential negative relation instances. With this heuristic in place, we only consider sentences with entity distance less than or equal to threshold parameter $\tau$.
Figure \ref{fig:analysis-see-dev} shows the impact of different values of $\tau$ on system performance. The value of $\tau \leq 20$ gives significant boost in precision with minor decrease in recall.  Our submission employed the threshold value of $\tau \leq 20$ between entity tokens.
\label{section:analysis-on-dev}

\subsection{Comparison with Participating Systems}

\textbf{SeeDev:} Table \ref{tab:SeedevBBTestScores} (left) is the official result of SeeDev Shared Task. Our submission {\it MIC-CIS} achieves the best score among all participating systems with F1 score of {\bf 0.373} showing compelling advantage. The system attains the highest precision ($0.294$) and recall ($0.511$). Precision and recall are not balanced however, and our system need an improvement to bring down false positives.

\begin{table}[t]
\center
\small
\renewcommand*{\arraystretch}{1.25}
\resizebox{.497\textwidth}{!}{
\setlength\tabcolsep{3.pt}
\begin{tabular}{r|c|r|c}
 \multicolumn{2}{c|}{{\bf Task: SeeDev}}					&   \multicolumn{2}{c}{{\bf Task: BB-norm+NER}} \\ \hline
\multicolumn{1}{c|}{\bf Team}  		&    $P$ \ / \    $R$ \  / \   $F1$     			        &  \multicolumn{1}{c|}{\bf Team}      &     $P$ \ / \    $R$ \  / \   $SER$   \\ \hline
{\bf MIC-CIS}						& {\bf .294} / {\bf .511} / {\bf .373}			&   {\bf MIC-CIS-1}       &       {\bf .624} / .433 / {\bf .715} \\
{\it YNU-junyi-1}					& .272 / .458 / .341		&   {\it MIC-CIS-2}       &       .560 / .449 / .786 \\
{\it Yunnan\_University-1}							&  .045 / .132 / .067   	&   {\it BLAIR\_GMU-1}       &         .496 / {\bf .467} / .793 \\
{\it Yunnan\_University-2}				       &     .020 / .132 / .035     	&   {\it BLAIR\_GMU-2}       &             .499 / .466 / .805 \\
{\it YNUBY-1}					       &  .011 / .070 / .019		&   {\it baseline-1}       &       .572 / .327 / .823 
\end{tabular}}
\caption{Comparison of our system (MIC-CIS) with top-5 participants: Scores on Test set for SeeDev and BB-norm+NER}
\label{tab:SeedevBBTestScores}
\end{table}

\textbf{BB-norm+NER:} Table \ref{tab:SeedevBBTestScores} (right) shows the comparison of performance among participating teams on BB-norm+NER test set. Our two submissions (MIC-CIS-1, MIC-CIS-2) ranked first and second with standard  error rate (SER) of \textbf{0.7159} and \emph{0.7867} respectively. The second submission employed Level2 NER to extract nested entities and hence has higher recall but with reduced precision. {\it MIC-CIS-1} has the highest precision 0.6242 and {\it MIC-CIS-2} has the recall close to the best recall of \emph{BLAIR\_GMU-1} with score \emph{0.4676}. Precision and recall are not balanced, we hypothesize improvement in nested entities extraction and modelling discontinuous entities will improve the system recall.

\section{Conclusion and Future Work}

In this paper, we described our system with which we participate in 
{\it PharmaCoNER}, {\it BB-norm+NER} and {\it SeeDev} shared tasks.
Our NER system employed linguistic features, multi-tasking via auxiliary objectives and hybrid loss including ranking loss to extract flat and nested entities in English and Spanish text. Our RE system employed SVM with linguistic features.
Compared to other participating systems, our submissions are ranked $1^{st}$ in BB-norm+NER and SeeDev task. Our system demonstrates competitive performance on PharmaCoNER with F1-score of $0.8662$.

In future, we would like to explore improved modelling strategies for nested NER and discontinuous entities extraction. Further, in this work we only addressed intra-sentence RE, we would be interested to explore approaches for inter-sentence RE \cite{DBLP:journals/tacl/PengPQTY17, DBLP:conf/aaai/GuptaRSR19}. Moreover, we would like to investigate interpretability of LSTMs for NER and RE \cite{DBLP:conf/emnlp/GuptaS18}.

\section*{Acknowledgment}
This research was supported by Bundeswirtschaftsministerium (bmwi.de), grant 01MD19003E (PLASS,  plass.io) at Siemens AG - CT Machine Intelligence, Munich Germany.

\bibliography{emnlp-ijcnlp-2019}
\bibliographystyle{acl_natbib}

\end{document}